\documentclass[runningheads]{llncs}
\usepackage{graphicx}

\usepackage{booktabs}

\begin{document}
\title{Using CollGram to Compare Formulaic Language in Human and Neural Machine Translation\thanks{The author is a Research Associate of the Fonds de la Recherche Scientifique - FNRS (Federation Wallonie Bruxelles de Belgique). He would like to warmly thank S. Granger and M. Dupont for access to the PLECI corpus.}}
\titlerunning{Comparing formulaic language in human and machine translation}
\author{Yves Bestgen\inst{1}\orcidID{0000-0001-7407-7797}}
\authorrunning{Y. Bestgen}
\institute{Universite catholique de Louvain, 10 place Cardinal Mercier, Louvain-la-Neuve, 1348, Belgium
\email{yves.bestgen@uclouvain.be}}
\maketitle              %
\begin{abstract}
A comparison of formulaic sequences in human and neural machine translation of quality newspaper articles shows that neural machine translations contain less lower-frequency, but strongly-associated formulaic sequences, and more high-frequency formulaic sequences. These differences were statistically significant and the effect sizes were almost always medium or large. These observations can be related to the differences between second language learners of various levels and between translated and untranslated texts. The comparison between the neural machine translation systems indicates that some systems produce more formulaic sequences of both types than other systems. 

\end{abstract}
\section{Introduction}
Neural machine translation (NMT) systems are currently considered to bridge the gap between human and machine translation \cite{Popel2020,Google}. However, little research has been done to determine whether NMT systems are also very effective in processing multiword units \cite{MON18,ZAN20}, whereas the importance of preformed units in language use is now well established, including in foreign language learning and translation \cite{Baker07,PAW83,SIN91}. The present study addresses this issue by comparing formulaic language in human and neural machine translation. It focuses on a specific category of multiword units, the "habitually occurring lexical combinations" \cite{Laufer11}, such as \textit{dramatic increase}, \textit{depend on}, \textit{out of}, which are not necessarily semantically non-compositional, but are considered statistically typical of the language because they occur "with markedly high frequency, relative to the component words or alternative phrasings of the same expression" \cite{BAL10}. These formulaic sequences (FSs) are analyzed by means of a technique proposed by \cite{Berna07}, improved by \cite{Dur09} and automated by \cite{BG14} under the name \textit{CollGram}\footnote{As one reviewer pointed out, this automation is not an "easily available plug-and-play implementation". However, there is a freely available system that implements it (http://collgram.pja.edu.pl) \cite{LEN16,WOL17}. Some of the indices used can also be easily obtained with the TAALES software \cite{KYL18} which allows the automatic analysis of many other lexical indices. TAALES presents however an important limitation because it only takes into account bigrams that occur at least 51 times in the reference corpus \cite{BE19rfla}, a value much too high for the MI at the heart of the CollGram approach.}. 

CollGram rests on two lexical association indices that measure the strength of attraction between the words that compose a bigram, mutual information (MI) and t-score \cite{EVE09}, calculated on the basis of the frequencies of occurrences in a reference corpus \cite{BG14,Dur09}. These two indices are complementary, MI favoring lower-frequency, but strongly-associated, FSs such as \textit{self-fulfilling prophecy}, \textit{sparsely populated} or \textit{sunnier climes} while the t-score favors high-frequency bigrams such as \textit{you know}, \textit{out of} or \textit{more than}. A series of studies have shown that, compared to native speakers, English as a foreign language learners tend to underuse collocations with high MI scores, while overusing those with high t-scores and that exactly the same differences are observed between advanced learners and intermediate learners \cite{BG14,Dur09}. These observations are in agreement with usage-based models of language learning which "hold that a major determining force in the acquisition of formulas is the frequency of occurrence and co-occurrence of linguistic forms in the input" \cite{Dur09}. It is worth noting that the same differences were observed between translated and untranslated texts, but the proposed explanation relies on a tendency towards normalization in translation \cite{Nancy,ConfTrad}. Since neural models also seem to be affected by frequency of use \cite{Koehn17,Li2020}, the hypothesis tested in the present study is that the same effects could be observed when comparing human translations (HTs) and NMTs, namely that NMTs will underuse high MI FSs and overuse high t-score FSs. 

\section{Method}
\subsection{Translation Corpus}
The texts used are taken from the journalistic section of the PLECI corpus (uclouvain.be/en/research-institutes/ilc/cecl/pleci.html). It is a sentence-aligned translation corpus of quality newspaper articles written in French and published in \textit{Le Monde diplomatique} and in English in one of the international editions of this same newspaper. Two hundred and seventy-nine texts, published between 2005 and 2012, were used for a total of 570,000 words in the original version and of 500,000 words in the translation. 

All original texts were translated into English by three well-known NMT systems: DeepL (deepl.com/translator), Google Translate (translate.google.com) and Microsoft Translator (microsoft.com/translator). Online translators were used for the first two, while the version available in \textit{Office 365} was used for the third. All these translations were performed between March 24 and April 6, 2021.

\subsection{Procedure}
Each translated text was tokenized and POS-tagged by CLAWS7 \cite{Ray03} and all bigrams were extracted. Punctuation marks and any character sequences that did not correspond to a word interrupted the bigram extraction. Each bigram, which did not include any proper name or number according to CLAWS, was then searched for in the 100 million word British National Corpus (BNC\footnote{\cite{Bes16Tal} showed that CollGram produces the same results if another reference corpus, such as COCA (corpus.byu.edu/coca) or WaCKy \cite{BAR09}, is used.}, www.natcorp.ox.ac.uk). When it is present, the corresponding MI and t-score were used to decide whether it is highly collocational or not. Based on \cite{ConfTrad} and \cite{Dur09}, bigrams with a score greater than or equal to 5 for the MI and 6 for the t-score were considered highly collocational. The last step consisted in calculating, for each text and for each association index, the percentage that the bigrams considered as highly collocational represent compared to the total number of bigrams present in the text.

\section{Analyses and Results}
Table 1 shows the average percentages of highly collocational bigrams for the MI and t-score in the four type of translations. The four means for each measure of association were analyzed using the Student's test for repeated measures since the same texts, which are the unit of analysis, were translated by the four translators. All these comparisons were statistically significant ($p < 0.0001$). 

\begin{table}
\begin{center}
\caption{Average percentages of highly collocational bigrams for the two indices in the four translation types.}
\begin{tabular}{|l|r|r|r|r|}
\hline
 Measure   &        Human  &    DeepL    &  Google   &   Microsoft \\ \hline
 High MI  &   11.21 & 10.48 & 10.07 & 10.27 \\
 High t-score    &  58.76 & 60.60 & 59.49 & 59.89 \\ \hline
\end{tabular}
\end{center}
\end{table}

Table 2 presents the differences between the means as well as two effect sizes. The first is Cohen's $d$, which expresses the size of the difference between the two means as a function of the score variability. According to \cite{Cohen88}, a $d$ of 0.50 indicates a medium effect and that a $d$ of 0.80 a large effect. The second effect size indicates the percentage of texts for which the difference between the two translations has the same sign as the mean difference. A value of 100 means that all texts produced by a given translator have a higher score than those translated by the other one and a value of 50 means that there is no difference.

\begin{figure}
 \begin{center}
  \includegraphics[width=11cm]{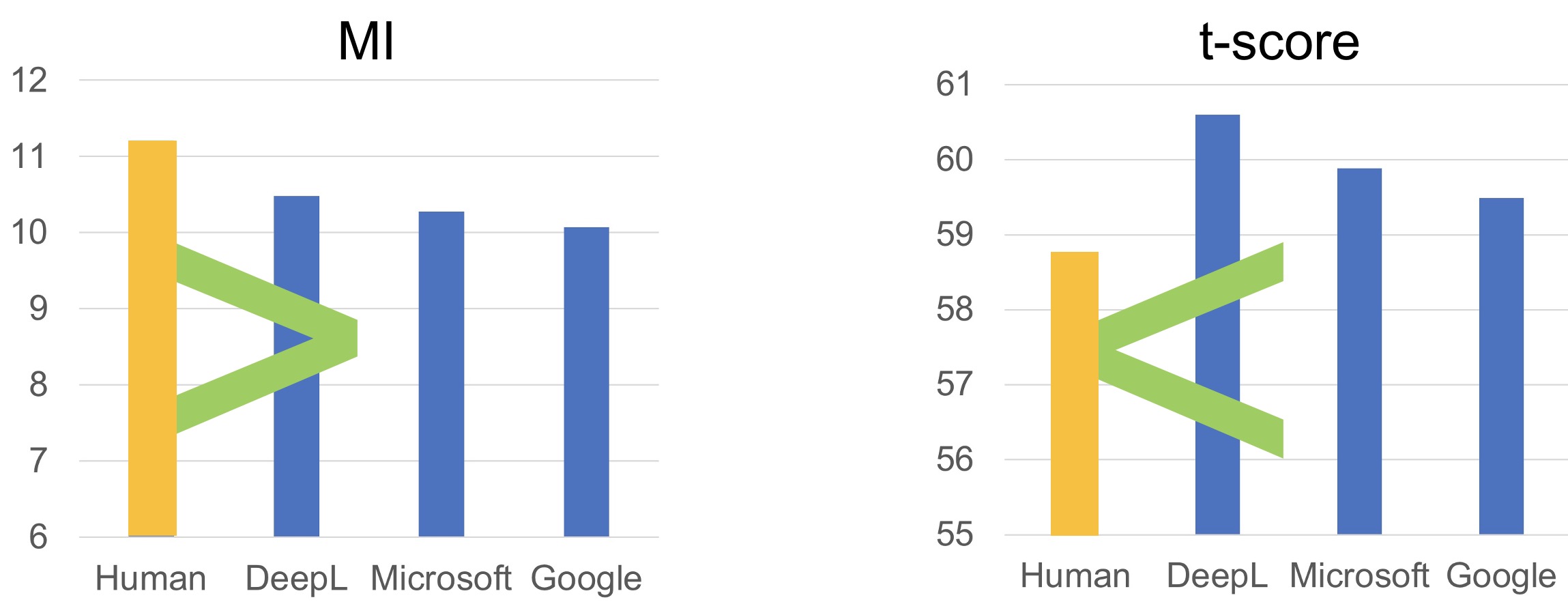}
 \end{center}
 \caption{Average percentages of highly collocational bigrams for the two indices.}
\end{figure}

As shown in these tables and in Figure 1, both hypotheses are verified. Compared to HTs, texts translated by the three neural systems contain a significantly smaller percentage of highly collocational bigrams for the MI and a larger percentage of highly collocational bigrams for the t-score. Cohen's $d$s are almost always medium or large and the percentages of texts for which differences are observed are greater than 70\% except in one case.

\begin{table}
\begin{center}
\caption{Differences (row translator minus column translator) and effect sizes for the two indices in the four translation types.}
\begin{tabular}{|l|rrr@{\hspace{0.5em}}|rrr@{\hspace{0.5em}}|rrr@{\hspace{0.5em}}|}
\hline
     &     \multicolumn{3}{|c}{Human} & \multicolumn{3}{|c|}{DeepL} & \multicolumn{3}{|c|}{Google} \\
   &   \multicolumn{1}{|c}{D} & \multicolumn{1}{c}{Es}  & \multicolumn{1}{c|}{\%}   &  \multicolumn{1}{|c}{D} & \multicolumn{1}{c}{Es}  & \multicolumn{1}{c|}{\%}  &   \multicolumn{1}{|c}{D} & \multicolumn{1}{c}{Es}  & \multicolumn{1}{c|}{\%} \\  \hline
& \multicolumn{9}{|c|}{High MI} \\
  DeepL & -0.72 & 0.59 & 73.48 & & & & & & \\
  Google & -1.14 & 1.00 & 84.33 & -0.41 & 0.65 & 74.91 & & &    \\
  Microsoft & -0.94 & 0.77 & 81.00 & -0.21 & 0.35 & 62.72  & 0.20 & 0.36 & 64.52 \\  \hline
& \multicolumn{9}{|c|}{High t-score} \\
  DeepL  & 1.83 & 0.84 & 80.65 & & & & & & \\
  Google & 0.72 & 0.32 & 62.37 & -1.11 & 0.98 & 84.59 & & & \\
  Microsoft &  1.13 & 0.51 & 71.33 & -0.70 & 0.62 & 70.97 & -0.41 & 0.42 & 69.18 \\  \hline
\end{tabular}
\end{center}
\end{table}

An analysis of the passages in which the differences between HT and NMT are the largest suggests that the origin lies at least partially in the less literal nature of human translations (see Table 3 for an example). 

\begin{table}
\begin{center}
\caption{Example of the four translation types and percentages of highly collocational bigrams for the two indices.}
\begin{tabular}{|l|l|r@{\hspace{0.5em}}|r@{\hspace{0.5em}}|}
\hline
Type & Phrase & \%High MI & \%High t-score \\ \hline
Original & A raison de huit heures par jour & & \\
Human & In an eight-hour day & 67 & 33 \\
DeepL & At eight hours a day & 25 & 100 \\
Google \& Microsoft & At the rate of eight hours a day & 14 & 100 \\
\hline
\end{tabular}
\end{center}
\end{table}

The differences between the three NMT systems are smaller, but still statistically significant. However, they require a different interpretation. When NMTs are compared to HTs (see Figure 1), the patterns of differences are reversed according to the MI or the t-score, as expected. For the NMT systems (see Figure 2), these patterns are identical for both types of collocation. The average percentages of highly collocational bigrams (see Table 1) are always higher in texts translated by DeepL than in those translated by Microsoft and also higher in the latter than in those translated by Google. Only a detailed qualitative analysis could determine whether these results indicate a difference in effectiveness.

\begin{figure}
 \begin{center}
  \includegraphics[width=11cm]{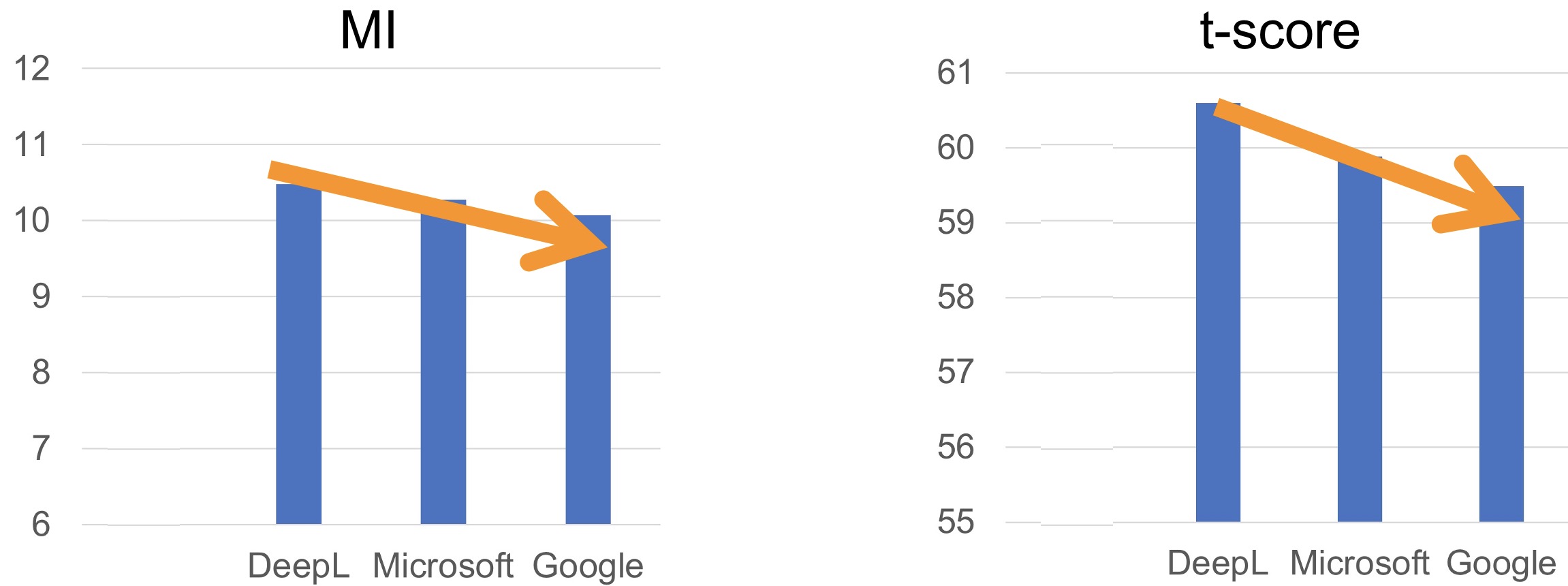}
 \end{center}
 \caption{Average percentages of highly collocational bigrams for the three NMT systems.}
\end{figure}

\section{Discussion and Conclusion}
The reported analyses confirm the hypotheses and thus suggest that, compared to HTs, NMTs more closely resemble texts written by intermediate learners than by advanced learners of English as a foreign language, a result that could be interpreted in the context of a usage-based model of language learning \cite{Dur09}. The NMTs also resemble translated texts more than untranslated texts, but it is not clear that this can be explained by a normalization process. Statistically significant differences, but smaller in terms of effect size, were also observed between the three NMT systems.

It is important to keep in mind that the present study only considers global quantitative properties of MWUs. At no point is the appropriateness in context of the MWUs assessed. It is therefore a very partial approach. However, it has the advantage of not requiring a human qualitative evaluation that is often complicated and cumbersome to set up. Moreover, it is likely that the appropriateness of a MWU is much more important for non-compositional expressions than for the habitually occurring lexical combinations studied here \cite{CON17}.

Another important feature of the approach is that it relies on a native reference corpus to identify highly collocational bigrams for both indices. As already mentioned, research on foreign language learning, but also on the comparison of translated and untranslated texts, has shown that the use of other large reference corpora such as COCA or WaCKy \cite{BAR09} did not change the results \cite{Nancy,Bes16Tal}. One can also wonder whether the use of a comparable reference corpus, rather than a generic one, would have returned different results. In the case of the comparison of translated and untranslated texts, \cite{Nancy} observed that the use of a journalistic corpus, the \textit{Corpus Est Republicain} (115 million words) made available by the Centre National de Ressources Textuelles et Lexicales, produced differences similar to those obtained with the WaCKy corpus. 

Before considering taking advantage of these observations to try to improve NMT systems, a series of complementary analyses must be conducted. Indeed, this study has many limitations, such as focusing only on a subcategory of MWUs \cite{MON18}, on a single language pair, and on a single genre of texts. Moreover, a thorough qualitative analysis is essential to better understand the results and evaluate the proposed explanations. As it has been shown in foreign language learning \cite{Bes17}, it would also be interesting to verify that the observed effects are not explained by differences in single-word lexical richness. Finally, the differences between the three NMT systems also require further analysis.

\bibliographystyle{splncs04}
\bibliography{Triton}

\end{document}